# *Automatic segmentation of lizard spots using an active contour model*

(Segmentación automática de manchas en lagartos usando un modelo de contornos activos)

*Authors:*

-Jhony Heriberto Giraldo Zuluaga[1], email: heriberto.giraldo@udea.edu.co.

-Augusto Enrique Salazar Jiménez[2, 3], email: augusto.salazar@udea.edu.co.

-Afilliation:

[1] Grupo de investigación GEPAR, Facultad de Ingenierías, Universidad de Antioquia UdeA, Calle 70 No. 52-21, Medellín, Colombia.

[2] Departamento de Ingeniería Electrónica y de Telecomunicaciones, Facultad de Ingeniería, Universidad de Antioquia UdeA, Calle 70 No. 52-21, Medellín, Colombia.

[3] Grupo de investigación AEyCC, Facultad de Ingenierías, Instituto Tecnológico Metropolitano ITM, Carrera 21 No. 54-10, Medellín, Colombia.

*Abstract:*

Animal biometrics is a challenging task. In the literature, many algorithms have been used, e.g. penguin chest recognition, elephant ears recognition and leopard stripes pattern recognition, but to use technology to a large extent in this area of research, still a lot of work has to be done. One important target in animal biometrics is to automate the segmentation process, so in this paper we propose a segmentation algorithm for extracting the spots of *Diploglossus millepunctatus,* an endangered lizard species. The automatic segmentation is achieved with a combination of preprocessing, active contours and morphology. The parameters of each stage of the segmentation algorithm are found using an optimization procedure, which is guided by the ground truth. The results show that automatic segmentation of spots is possible. A 78.37 % of correct segmentation in average is reached.

*Resumen:*

La biometría en animales es una tarea desafiante. En la literatura muchos algoritmos se han utilizado, como reconocimiento de los pechos en pingüinos, reconocimiento de las orejas en elefantes y reconocimiento de los patrones de rayas en leopardos por ejemplo. Aún hay mucho trabajo para hacer un uso masivo de la tecnología. En este artículo proponemos un algoritmo de segmentación para extraer manchas de la especie de lagartos *Diploglossus millepunctatus.* Esta es una especie amenazada por la actividad humana. La segmentación automática ha sido lograda con una combinación de preprocesamiento, contornos activos y morfología. Los parámetros de cada etapa del algoritmo de segmentación han sido optimizados usando imágenes de referencia como objetivo. Los resultados muestran que la segmentación automática de manchas es posible. Un 78.37% de segmentación correcta en promedio es alcanzado.

Palabras clave: Diploglossus millepunctatus, contornos activos, corrección gamma, filtrado morfológico, segmentación de manchas.

1. *Introduction:*

Biometric identification in humans has been treated with digital signal processing aiming at recognizing hands, faces, voice, eyes, DNA, heart sound, ethnicity, scars, marks and tattoos [1]. Each topic has been extensively covered by techniques such as features extraction and machine learning. This investigation field is applied to surveillance and security systems, database extraction, tracking, and others.

The biometric methods of animal identification are application dependent, because each animal species has unique characteristics. We can search for characteristics to distinguish one individual from another one in the same species. In the literature, there are some examples, e.g. multi-curve matching of the elephants' ears is used for identifying elephants [2]. Another example is monitoring populations of penguins using their chest spots [3]. Animal biometrics is an emerging research field that combines pattern recognition, ecology and information sciences [4].

The biometric recognition of lizards is little reported in the literature, one example is the *Pygmy bluetongue* lizard. This lizard was identified by signature curves in [5]. The aforementioned dotted lizard (*Diploglossus millepunctatus*) is a lizard species native to Malpelo Island, located in the Colombian Pacific. The identification of this animal species is important, because it is in danger of extinction due to human activities [6]. To our knowledge, there is no non-invasive biometric identification method in the literature to distinguish individuals of *Diploglossus millepunctatus*. The method proposed here is non-invasive and based on image processing. Our algorithm includes segmentation and classification steps to identify each lizard. Various problems on spot identification can occur due to lighting variation [7] or problems due to perspective [8] that are discussed in this paper. The idea is to use segment spot patterns of *Diploglossus millepunctatus* to solve the named problems. In our method, a gamma correction algorithm was applied to get rid of lighting problems. Problems due to perspective were addressed by a deformable active contour model.

Some algorithms avoid the segmentation part using computer aided algorithm [7] [8] [9]; in this paper, we propose an automatic algorithm that solves the segmentation problem by dividing the segmentation into two threads. Firstly, the algorithm is applied to the image without preprocessing, with the aim to extract the darkest spots. The second thread is a gamma correction, and then the algorithm is applied to the image, with the aim to extract the lightest spots. Each thread consists of three stages. First, preprocessing that prepares the image for the active contours iteration. Second, active contours, which are not training but their parameters are tuned by an optimization algorithm; and third, deletion of atypical images via a morphological operation.

The paper is organized as follows. Section 2 shows related works. Section 3 describes the active contour method, the segmentation and the optimization algorithms. Section 4 presents the experimental framework. Section 5 exhibits the experimental results and Section 6 shows the conclusion and future works.

2. *Related works:*

Spots segmentation has been used for medical purposes. One of the most important applications is disease diagnosis [10]. For disease diagnosis some methods are found in the literature. [11] uses Bayesian classification combined with Markov Random Field. [12] uses a watershed segmentation algorithm and [13] an active contours algorithm. The principal target of each algorithm is similar to our proposed method, but those algorithms have been used for medical purposes. The algorithm presented in this paper is used for biological purposes. Our images have illumination problems and the image is deformed by the perspective, thus the results of the segmentation of images for medical purposes and our algorithm cannot be numerically comparable. Figure 1 shows the *Diploglossus millepunctatus.* As can be seen, the lizard's skin is covered by a natural oily substance that gives it an unavoidable shine. The idea of this paper is to extract some lizard scales and segment the spots that are on the scales. The lizard's scales represent sets of spots that are limited by grooves.

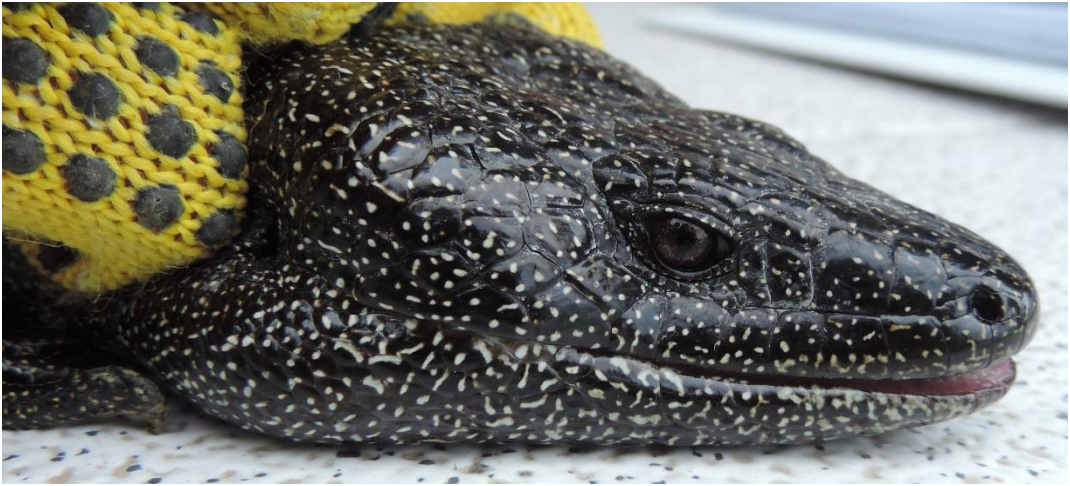

Figure (1). Diploglossus millepunctatus.

3. *Method:*

This section explains the methods used in this paper. First, the active contours method is explained, then the segmentation algorithm used is elucidated; finally, the optimization technique for finding the best parameters for the segmentation is explained.

3.1. *Active contours:*

An active contours or snake is an energy-minimizing spline, guided by external constraint forces and influenced by image forces that pull it towards features such as lines and edges [14]. Active contours take an initial condition that is automatic or set up by the user.

The essential idea of active contours is to take a feature map $F(r)$. The snake is a deformable curve $r(s)$, $0 \leq s \leq 1$ can slither on $F(r)$. Equilibrium equations, $r(s)$ tend to cling to high responses of $F(r)$. The equilibrium equation is shown in Equation 1. In this equation, the tendency to maximize $F(r)$ is formalized as the external energy. The counterbalance of the external energy is known as internal energy, which tends to preserve the smoothness of the curve [15]. The coefficients $\omega_1$ and $\omega_2$ in the equation 1, which must be positive, are the restoring forces associated with the elasticity and stiffness of the snake, respectively.

$$\left(\frac{\partial(\omega_1 r)}{\partial s} - \frac{\partial^2(\omega_2 r)}{\partial s^2}\right) + \nabla F = 0$$

Equation (1): Energy formulation

Active contours have the advantage that they do not need any previous training algorithm. Equation 1 was solved with the Chan-Vese method [16].

3.2. *Segmentation algorithm:*

The segmentation algorithm is divided into two threads. One of the threads consists of four stages; the other thread consists of five stages. The stages are gray conversion, median filter, gamma correction, active contours and area opening. This procedure is shown in Figure 2. One thread performs segmentation without gamma correction to extract spots in the dark region; the other thread performs segmentation with gamma correction to extract spots in the bright region.

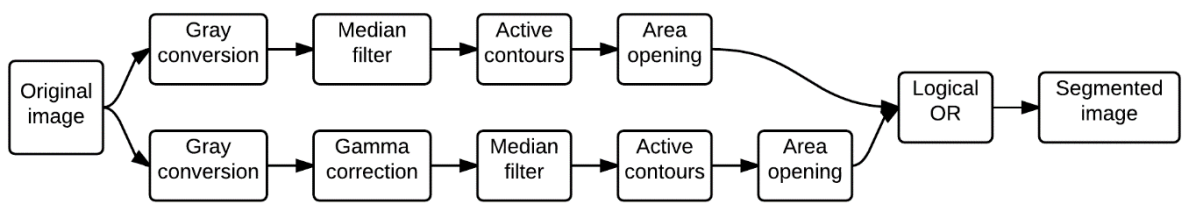

Figure (2): Block diagram, segmentation algorithm.

The first thread (dark region) consists of a color space transformation that is a linear combination of the original space Red Green and Blue (RGB) to result in gray space, a median filter to homogenize the region, the principal active contour iterations, and finally the area opening for deleting atypical spots. In this section, the image is processed without luminance correction, the idea is to extract spots in the darkest regions.

The gray intensity value of each pixel from RGB to gray space is defined as a weighted sum of three linear values. The luminance $Y$ in terms of CIE 1931 is given by Equation 2 [17], where $R$, $G$ and $B$ are the level of red, green and blue, respectively.

$$Y = 2.986R + 0.5870G + 0.1140B$$

Equation (2): RGB to Luminance Y (CIE 1931)

The second thread (bright region) uses the same color space transformation as the first thread, but, additionally, a decoding gamma correction. In this stage, the image is processed with a nonlinearity operation to extract the spots that are in the bright regions.

Gamma correction is a non-linear operation called power-law. Equation 3 shows the power-law of the gamma correction [18], where $l \in [0,1]$ denotes the image pixel intensity, $c$ is a constant, in common cases $c = 1$, and $\gamma$ denotes the gamma constant. The value of $\gamma$ is found by optimization techniques.

$$S = cl^\gamma$$

Equation (3): Gamma correction

Let $\alpha$ be the percentage of the area opening $\alpha \in [0,1]$, the value of $\alpha$ is found by optimization techniques. Let $w$ and $h$ be the image dimensions and let $m_i$ be the area of the spot $i$. Equation 4 shows the area opening algorithm.

$$if\ m_i \geq \alpha wh\ then\ m_i = 0\ \forall i$$

Equation (4): Area opening algorithm

Figure 3 shows an example of the preprocessing stage. Figure 3.a shows one original image. Figure 3.b shows the first thread (dark region) of image preprocessing. As can be seen, the luminance problem is not resolved, because the idea is to extract the spots in the dark region. Figure 3.c shows the second thread (bright region) of image preprocessing. This figure shows that the spots in dark regions are overshadowed. However, the spots in bright regions are not overshadowed, so it is possible to extract the spots that are in bright regions with this preprocessing procedure.

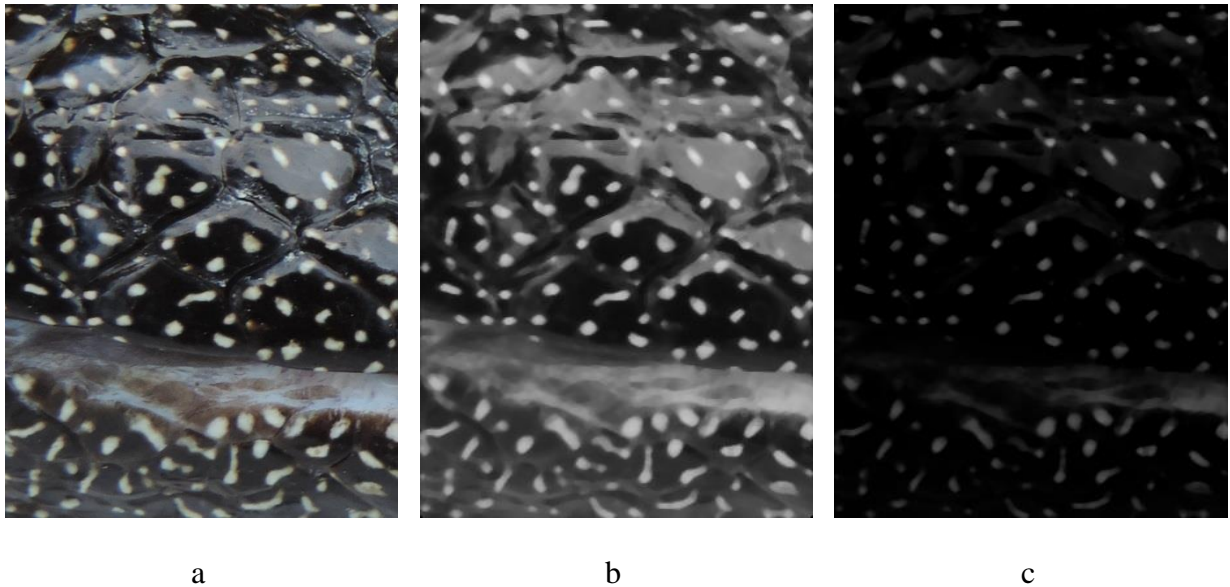

a  b  c

Figure (3). Preprocessing images. a) original image. b) preprocessed image without gamma correction. c) preprocessed image with gamma correction.

The result of active contours is binary images for each thread. Finally the two threads are merged by a logical *or*.

### 3.3. *Optimization algorithm:*

It is necessary to choose the best parameters for the segmentation algorithm. To reach this objective, an exhaustive search was run. First we established a ground truth of some lizard scales under various conditions.

In computer vision the ground truth (GT) plays an important role in the evaluation process. The GT is important for the development of new algorithms, to compare different algorithms, and to evaluate performance, accuracy and reliability [19]. For instance, in this paper, Figure 4.a shows one original image of the database and Figure 4.b shows its corresponding ground truth image. The ground truth validation is used in this paper to help the optimization algorithm that is explained below.

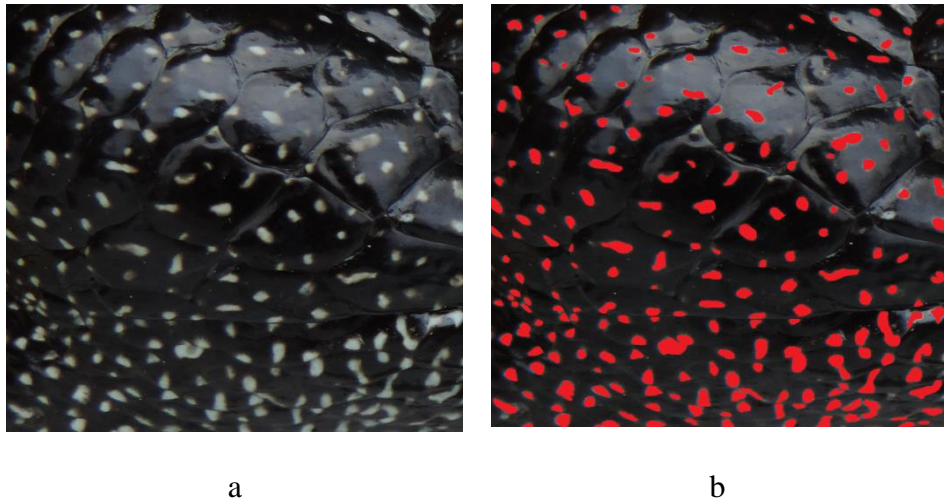

a  b

Figure (4). Ground truth process a) original image, b) processed image.

Let **X** be the 2x2 confusion matrix (background and foreground) between the ground truth image and the segmented image, $X_{11}$ is the percentage of the background that was segmented as background, $X_{12}$ is the percentage of the foreground that was segmented as background, $X_{21}$ is the percentage of background that was segmented as foreground and $X_{22}$ is the percentage of the foreground that was segmented as foreground.

For future work, the idea is to prevent the loss of information, which means that all spots should be present in the segmented image. The objective function of the optimization algorithm is shown in Equation 5. The aim is to maximize the percentage of correct segmentation. The confusion matrix was extracted comparing the ground truth image and the segmented image.

$$\text{Maximize}(X_{11} + X_{22})$$

Equation (5): Objective function

The restrictions of the algorithm are shown in Equation 6 where $\gamma$ is the gamma correction factor; $\rho$ is the iteration number in active contours and $\alpha$ is the percentage in the area opening. With this algorithm, the best parameters of the segmentation were found.

$$3.6 \leq \gamma \leq 6$$

$$1600 \leq \rho \leq 2600$$

$$0.05 \leq \alpha \leq 0.0025$$

$$\gamma, \alpha \in R \text{ and } \rho \in Z^+$$

Equation (6): Restriction

4. *Experimental framework:*

This section explains the database conformation and describes the executed experiments. First, the optimization algorithm was executed, followed by the segmentation algorithm.

4.1. *Database:*

The database consists of 20 samples (lizards). The images were taken under controlled conditions. 20 pieces of the original images were extracted to apply the optimization algorithm. These pieces of the original images have a resolution of 700x700 pixels, and basically are lizard scales with some spots. The database is limited, because the ground truth method is a demanding process. This is because an expert has to manually segment each image in the database.

There are three kinds of light exposition lizard scales: normal (images with normal conditions of luminance), ideal (images without luminance exposure), and hard exposed images (images with high exposures to light). Figure 5 shows each kind of light condition that is present in the database.

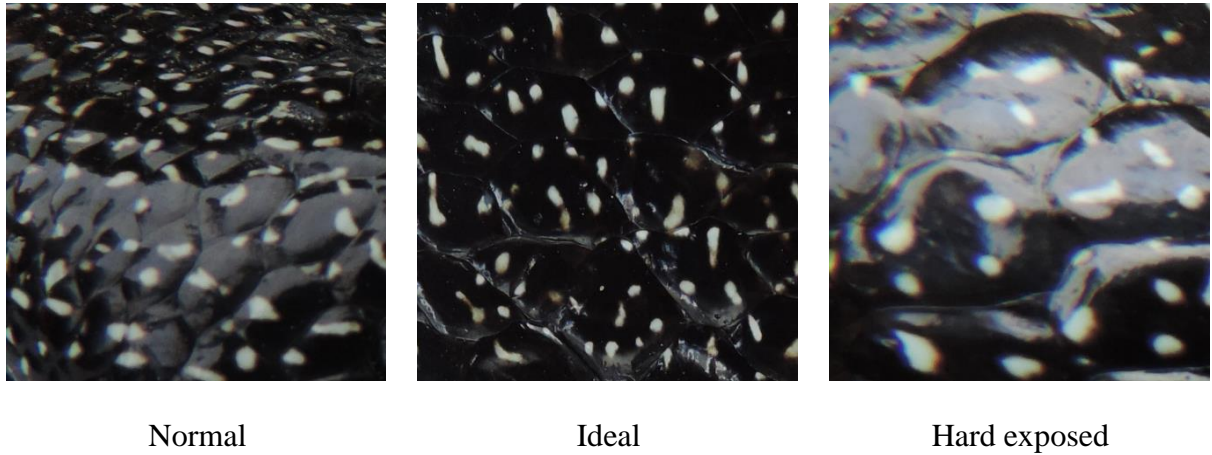

| Normal | Ideal | Hard exposed |

Figure (5): Different light conditions on lizard scales

### 4.2. *Testing:*

First, the optimization algorithm was run on 20 images with combinations of all restricted parameters. For each image, the segmented algorithm was applied with 546 combinations of parameters. The 546 combinations result from 13 selected discrete values of $\gamma$: six selected discrete values of $\rho$, and seven selected discrete values of $\alpha$. The discrete values were selected with the experience. For each combination of parameters, the confusion matrix was extracted and the best parameter for each image was saved. The best parameters were selected with respect to the objective function of the optimization algorithm. For the selected parameters, the median was applied.

The validation testing of the segmentation was run with the segmentation algorithm on 95 images of the database. The parameters of the segmentation algorithm were the parameters found by the optimization algorithm explained above. For each segmented image, the binary confusion matrix was extracted with respect to the ground truth. The median and standard deviation were calculated for all obtained confusion matrices.

### 5. *Experimental results:*

This section shows the experimental results of the segmentation algorithm, followed by a discussion about the objective function of the optimization algorithm. Finally, some visual results are exhibited.

Table 1 shows the result of the segmentation algorithm. The performance of the segmentation algorithm is 78.37% that is $(X_{11} + X_{22})/2$.

|  | *Background* | *Foreground* |
|---|---|---|
| *Background* | 96.58% ± 1.94% | 39.84% ± 22.9% |
| *Foreground* | 3.42% ± 1.94% | 60.16% ± 22.9% |

Table (1): Confusion matrix results

Table 2 shows the classical metric precision, recall and f-measure [20]. Each metric was extracted with the images of the validation set, and the mean and standard deviation are shown in the table.

| *Metric* | *Precision* | *Recall* | *F-measure* |
|---|---|---|---|
| *Value* | 60.80% ± 22.10% | 45.53% ± 14.48% | 49.79% ± 15.32% |

Table (2): Classical metrics

The results of the segmentation algorithm in the validation have high standard deviation results. The classical metrics support this affirmation, where the standard deviation of each metric is close to 20%. The precision metric gives us a notion of correct spots over all supposed spots in the segmentation algorithm. The results of the classical metrics have high standard deviations, maybe due to the variability of the light exposure.

Figure 6 shows some visual results of the segmentation algorithm. Figure 6.a shows the original image. Figure 6.b shows the binary ground truth images. Figure 6.c shows the segmented images; and Figure 6.d shows the comparative image algorithm. Red-colored regions mean spots that are on the ground truth image but not on the segmented image (false negatives). Spots with yellow color mean spots that are in both, the ground truth and segmented images (true positives). Green color spots mean spots that are in the segmented image but not in the ground truth image (false positives).

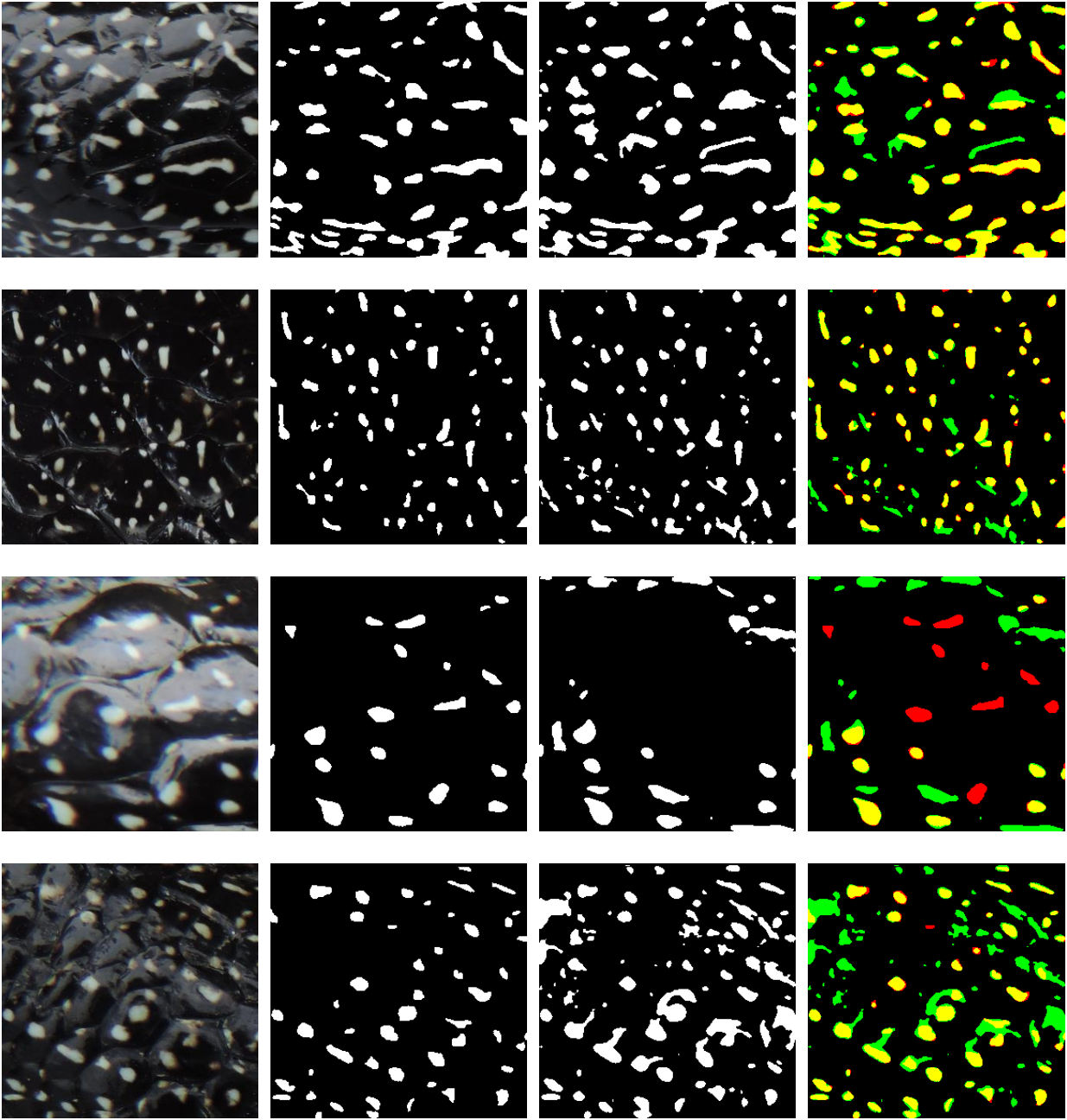

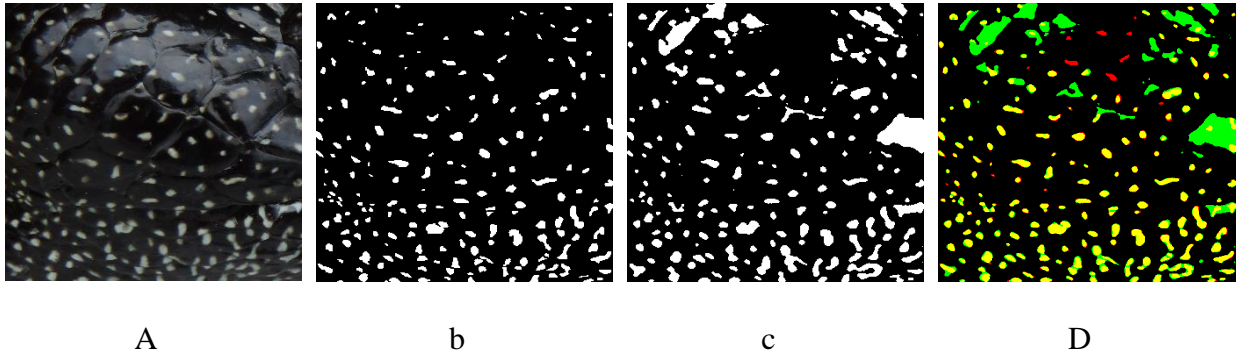

| A | b | c | D |

Figure (6): Some results of the optimization and segmentation algorithm. a) original image b) ground truth image, c) segmented image d) comparative image.

Figure 6.d shows that the segmentation algorithm has some limitations, especially with high expose images. The limitations could be caused, because the search space shown in Equation 6 is a reduced search space, which means that the solution given by the optimization algorithm could be a local minimum. In future work, this problem can be solved using a heuristic technique to avoid local minima.

Finally, the average running time of the segmentation algorithm was $143.9 \pm 53.3$ seconds per each segmented image. The experiments were carried on an Intel Core i7 3630QM with 8 gigabytes of RAM memory.

6. *Conclusion and future work:*

We introduced an algorithm for automated segmentation of lizard spots of *Diploglossus millepunctatus*. The algorithm is composed of a preprocessing stage, active contours iterations and morphological filtering. The best parameters for the three algorithm stages were selected using an optimization with the objective to segment right spots and right background. The optimization algorithm needs a better design in view of practical applications, e.g. with other animals, because the processing time is already quite high. Another important aspect is that the algorithm execution, after optimization, is not as time-demanding as other segmentation algorithms in the state of the art, e.g. Markov Random Field, graph cuts and so on.

For future work, a heuristic technique will reduce the training time problem and limitations present in Equation 6, because the optimization algorithm will be trained with more images to attain a better and more intelligent search. The next phase is the identification stage, to identify, whether a certain scale belongs to a certain lizard. In the identification stage, Procrustes Analysis, and registration algorithms will be tested.